\documentclass[sigconf]{acmart}
\renewcommand\footnotetextcopyrightpermission[1]{}
\usepackage{multirow}
\usepackage{colortbl}
\usepackage{enumitem}

\AtBeginDocument{%
  }

\setcopyright{acmlicensed}
\copyrightyear{2018}
\acmYear{2018}
\acmDOI{XXXXXXX.XXXXXXX}
\acmConference[Conference acronym 'XX]{Make sure to enter the correct
  conference title from your rights confirmation email}{June 03--05,
  2018}{Woodstock, NY}
\acmISBN{978-1-4503-XXXX-X/2018/06}

\begin{document}

\title[MM-DADM for Virtual Clinical Trials]{MM-DADM: Multimodal Drug-Aware Diffusion Model for\\Virtual Clinical Trials}

\author{Qian Shao}
\authornote{Both authors contributed equally to this research.}
\email{12221112@zju.edu.cn}
\orcid{0000-0001-5768-6136}
\affiliation{%
  \institution{Zhejiang University}
  \city{Hangzhou}
  \state{Zhejiang}
  \country{China}
}

\author{Bang Du}
\authornotemark[1]
\email{bangdu1994@zju.edu.cn}
\affiliation{%
  \institution{Zhejiang University}
  \city{Hangzhou}
  \state{Zhejiang}
  \country{China}
}

\author{Zepeng Li}
\affiliation{%
  \institution{Zhejiang University}
  \city{Hangzhou}
  \state{Zhejiang}
  \country{China}
}

\author{Qiyuan Chen}
\affiliation{%
  \institution{Zhejiang University}
  \city{Hangzhou}
  \state{Zhejiang}
  \country{China}
}

\author{Jiahe Chen}
\affiliation{%
  \institution{Zhejiang University}
  \city{Hangzhou}
  \state{Zhejiang}
  \country{China}
}

\author{Hongxia Xu}
\affiliation{%
  \institution{Zhejiang University}
  \city{Hangzhou}
  \state{Zhejiang}
  \country{China}
}

\author{Jimeng Sun}
\affiliation{%
  \institution{University of Illinois Urbana-Champaign}
  \city{Champaign}
  \state{Illinois}
  \country{USA}
}

\author{Jian Wu}
\authornote{Corresponding authors.}
\email{wujian2000@zju.edu.cn}
\affiliation{%
  \institution{Zhejiang University}
  \city{Hangzhou}
  \state{Zhejiang}
  \country{China}
}

\author{Jintai Chen}
\authornotemark[2]
\email{jtchen721@gmail.com}
\affiliation{%
  \institution{The Hong Kong University of Science and Technology (Guangzhou)}
  \city{Guangzhou}
  \state{Guangdong}
  \country{China}
}

\renewcommand{\shortauthors}{Trovato et al.}

\begin{abstract}
High failure rates in cardiac drug development necessitate virtual clinical trials via electrocardiogram (ECG) generation to reduce risks and costs.
However, existing ECG generation models struggle to balance morphological realism with pathological flexibility, fail to disentangle demographics from genuine drug effects, and are severely bottlenecked by early-phase data scarcity.
To overcome these hurdles, we propose the Multimodal Drug-Aware Diffusion Model (MM-DADM), the first generative framework for generating individualized drug-induced ECGs.
Specifically, our proposed MM-DADM integrates a Dynamic Cross-Attention (DCA) module that adaptively fuses External Physical Knowledge (EPK) to preserve morphological realism while avoiding the suppression of complex pathological nuances.
To resolve feature entanglement, a Causal Feature Encoder (CFE) actively filters out demographic noise to extract pure pharmacological representations. These representations subsequently guide a Causal-Disentangled ControlNet (CDC-Net), which leverages counterfactual data augmentation to explicitly learn intrinsic pharmacological mechanisms despite limited clinical data.
Extensive experiments on $9,443$ ECGs across $8$ drug regimens demonstrate that MM-DADM outperforms $10$ state-of-the-art ECG generation models, improving simulation accuracy by at least $6.13\%$ and recall by $5.89\%$, while providing highly effective data augmentation for downstream classification tasks.
\end{abstract}



\keywords{multimodal fusion, diffusion model, electrocardiogram generation, virtual clinical trials}


\maketitle

\section{Introduction}
The development of novel cardiac drugs is fraught with challenges, marked by exorbitant clinical trial costs and high attrition rates~\citep{fordyce2015cardiovascular, dimasi2016innovation}.
A staggering $79\%$ of failures in late-stage trials are attributed to insufficient efficacy or unforeseen safety issues, underscoring a critical gap in early-stage predictive capabilities~\citep{dowden2019trends}.
Therefore, conducting virtual clinical trials by simulating drug-induced ECG changes has become a powerful strategy to reduce risk and cost.

\begin{table}[tbp]
\caption{Previous ECG GMs vs. MM-DADM.}
\centering
\begin{tabular}{l|cc}
\toprule
\multicolumn{1}{l|}{Property}
& \multicolumn{1}{c}{Previous ECG GMs}
& \multicolumn{1}{c}{MM-DADM (Ours)}   \\
\cmidrule{1-3}
Requires data
& \textcolor[RGB]{204,0,0}{Large quantities}   & \textcolor[RGB]{0,204,0}{Small quantities} \\
Integrates EPK
& \textcolor[RGB]{204,0,0}{Static, rigid}      & \textcolor[RGB]{0,204,0}{Dynamic, adaptive} \\
Disentangles info
& \textcolor[RGB]{204,0,0}{No}                 & \textcolor[RGB]{0,204,0}{Causal-disentangle} \\
\bottomrule
\end{tabular}
\label{t1}
\end{table}

However, adapting existing ECG generation models (GMs)~\citep{simone2023ecgan,chung2023text,li2024biodiffusion} for high-fidelity pharmacodynamic modeling presents three challenges (Table~\ref{t1}).
First, balancing morphological realism with pathological flexibility remains a fundamental hurdle. While previous studies employ Ordinary Differential Equations (ODEs) as External Physical Knowledge (EPK) to enforce strict physiological rules~\citep{mcsharry2003dynamical,golany2020simgans,golany2021ecg}, these rigid constraints inherently suppress the complex, highly variable morphological deviations induced by specific drugs.
Second, generating individualized drug responses, while ensuring authenticity, remains largely unexplored. We find that current conditional GMs fail to disentangle demographics from drug-related information~\citep{golany2019pgans,li2024biodiffusion,neifar2024leveraging}, substantially limiting their ability to simulate pharmacological effects (see Section~\ref{pcds}). The simulated effects may be driven by spurious demographic correlations rather than intrinsic pharmacological mechanisms. This flaw is fatal: models may mask genuine drug-induced cardiotoxicity under healthy demographic profiles, thereby misleading drug screening.
Finally, and most critically, the resolution of this feature entanglement problem is severely constrained by the inherent scarcity of clinical data. Unlike prior ECG generation tasks, which aim to augment data for downstream disease diagnosis and typically assume sufficient training data~\citep{chen2022me,hu2024personalized,yang2024data}, our goal is to simulate drug responses using strictly limited early-phase clinical trial data, thereby reducing the need for large-scale physical trials. In other words, if abundant pre- and post-dose ECG data were already available, synthesizing such data would lose its clinical value.

To address these limitations, we propose the Multimodal Drug-Aware Diffusion Model (MM-DADM), a novel generative framework to simulate individualized drug-induced ECGs.
For the first challenge, we introduce a Dynamic Cross-Attention (DCA) module.
Instead of strictly enforcing ODE-based physical priors, DCA adaptively fuses this EPK into the diffusion process based on the semantic discrepancy between the EPK and the denoised latent feature. This preserves basic morphological realism without suppressing complex pathological nuances.
For the second one, we pre-train a causal feature encoder (CFE) using contrastive learning. This module actively filters out patient-specific noise to extract pure drug-specific representations, maintaining a dynamic prototype that serves as the canonical representation of each drug's unique causal effect.
For the third one, we design the Causal-Disentangled ControlNet (CDC-Net) equipped with a counterfactual data augmentation strategy.
By synthetically recombining demographic variables with unmatched drug variables, we generate expansive counterfactual samples. Guided by the pre-trained drug prototypes, CDC-Net is explicitly forced to learn intrinsic pharmacological mechanisms, ensuring the synthesized ECG variations are genuinely driven by the assigned drug.

We compare MM-DADM with $10$ state-of-the-art (SOTA) generative models on $2$ public datasets, encompassing $9,443$ $12$-lead ECGs from $8$ different drug regimens.
The quantitative analyses and expert assessments demonstrate that MM-DADM generates ECGs with superior fidelity.
The comparative results verify that MM-DADM can more accurately simulate drug-induced changes in ECGs, improving the accuracy by at least $6.13\%$ and recall by $5.89\%$.
Furthermore, we show that synthetic ECGs from MM-DADM can serve as effective data augmentation for improving downstream drug-effect classification tasks.
Ablation studies and visualizations also confirm the efficacy of each proposed component.

In summary, our main contributions are as follows:
\begin{itemize}
    \item We propose MM-DADM, a multimodal fusion framework to generate high-fidelity, drug-induced ECGs. To our best knowledge, we are the first to simulate drug effects on ECGs to conduct virtual clinical trials.
    \item We introduce the DCA fusion to fuse EPK throughout generation, ensuring the authenticity of ECG waveforms.
    \item We design a CFE pre-trained via contrastive learning to actively filter out patient-specific demographic noise and extract pure, disentangled pharmacological representations.
    \item We develop the CDC-Net, leveraging learned drug prototypes to robustly simulate individualized ECGs.   
    \item We conduct comparison experiments to demonstrate MM-DADM's SOTA performance against $10$ competing models and its utility in downstream tasks.
\end{itemize}
\section{Related Work}


GMs have emerged as a pivotal technology for simulating physiological time series, particularly ECGs~\citep{rouzrokh2025current}. The growing reliance on GMs is primarily driven by systemic challenges in medical artificial intelligence: severe class imbalance due to the rarity of specific arrhythmias~\citep{berger2023generative}, ubiquitous data scarcity~\citep{rouzrokh2025current}, and stringent privacy regulations that impede patient data sharing~\citep{adib2025synthetic}.
To address these bottlenecks, a diverse array of generative architectures has been widely explored for ECG synthesis, including Generative Adversarial Networks (GANs)~\citep{simone2023ecgan}, Variational Autoencoders (VAEs)~\citep{chung2023text}, Diffusion Models (DMs)~\citep{li2024biodiffusion}, Flow Matching Models (FMMs)~\citep{bondar2025flowecg}, and Multimodal Foundation Models (MFMs)~\citep{chung2023text}.
Consequently, contemporary research predominantly leverages these models for two critical applications: (1) serving as an advanced data augmentation technique to synthesize rare pathological samples, thereby enhancing the robustness of downstream diagnostic classifiers~\citep{berger2023generative}, and (2) generating high-fidelity, privacy-preserving synthetic cohorts to facilitate unrestricted collaborative research~\citep{adib2025synthetic}.

\paragraph{GANs}
In the realm of ECG synthesis, GANs are the dominant research paradigm~\citep{venugopal2024boosting}.
However, the original GAN~\citep{goodfellow2020generative} suffers from training instability and long-range dependencies, which affect the authenticity of the generated results.
For the former, researchers adopted more robust variants such as Wasserstein GAN (WGAN)~\citep{arjovsky2017wasserstein}.
For the latter, research shifted to Transformer-based architectures like TTS-CGAN~\citep{li2022tts}, which employs self-attention mechanisms in both generator and discriminator.
With authenticity guaranteed, an increasing number of studies focus on generating personalized ECGs. For example, \citet{golany2020simgans,golany2021ecg} integrate physical knowledge into the generation processes of GANs to produce realistic ECG signals by using a physically motivated simulator, given by a set of ODEs~\citep{mcsharry2003dynamical}. \citet{golany2019pgans} uses the wave values and durations of ECG signals as constraints for GAN to generate personalized heartbeats. \citet{hu2024personalized} generates digital twins of healthy individuals’ anomalous ECGs and enhances the model's sensitivity to the personalized symptoms.

\paragraph{VAEs}
VAEs offer an alternative generative paradigm, characterized by stable training and the ability to learn a structured, probabilistic latent space~\citep{kingma2019introduction}. This structured representation makes them naturally suited for conditional generation.
For instance, \citet{beetz2022multi} proposes a multi-domain VAE, which integrates anatomy information. \citet{sang2025deep} proposes a cVAE to generate ECGs by incorporating demographic metadata and anatomical heart features. However, traditional VAEs are often criticized for producing overly smooth or blurry outputs, struggling to reproduce high-frequency details in ECGs~\citep{ vivekananthan2024comparative}.
Recent architectural innovations, such as cNVAE-ECG~\citep{sviridov2025conditional}, have mitigated this issue, achieving high-resolution conditional synthesis and demonstrating superior performance to GANs in downstream classification tasks.

\paragraph{DMs}
More recently, DMs have emerged as the SOTA in the generative field, which can generate samples with both high fidelity and diversity, effectively overcoming the instability of GANs and the blurriness of VAEs.
For instance, Diffecg~\citep{neifar2023diffecg} is designed for ECG generation, imputation, and forecasting.
SSSD-ECG~\citep{alcaraz2023diffusion} uses a structured state space model to generate short $12$-lead ECGs.
While DSAT-ECG~\citep{zama2023ecg} introduces a novel DM, focusing on confidentiality issues related to sensitive health data distribution.

\paragraph{FMMs}
Despite the impressive results of diffusion models, their iterative sampling process is computationally expensive.
This has spurred exploration into flow matching~\citep{lipman2023flow}, a related but more sample-efficient method, as demonstrated by FlowECG~\citep{bondar2025flowecg}, which maintains generation quality while significantly reducing the number of sampling steps.

\paragraph{MFMs}
A parallel and rapidly expanding frontier is the use of large-scale foundation models for ECG analysis. This paradigm shifts the objective from unconditional signal generation to multimodal reasoning.
For instance, \citet{chung2023text} proposes an autoregressive generation model Auto-TTE driven by clinical text reports.
More advanced models such as BioMedGPT~\citep{luo2023biomedgpt} and the specialized ECG-LM~\citep{yang2025ecg} align the signal feature space (from a specialized ECG encoder) with the text feature space of a large language model.
Their task is no longer signal synthesis, but rather diagnostic reasoning, enabling complex applications like ECG Question-Answering and automated clinical report generation.

Most of the above works have made significant progress. However, they have mainly focused on auxiliary diagnosis of diseases. In contrast, our work is centered on generating ECGs that reflect drug reactions, thereby enabling virtual clinical trials.
\section{Methods}

In this section, we first review the ODE system to derive the EPK and the denoising diffusion backbone in Section~\ref{odes} and Section~\ref{ddpm}, respectively.
To effectively integrate this physical prior, we introduce a DCA module that adaptively fuses the EPK into the denoising process in Section~\ref{dcaf}.
Subsequently, to disentangle pharmacological effects from demographics, we pre-train a CFE in Section~\ref{ept}.
Finally, we detail the CDC-Net to ensure the generated ECGs are intrinsically driven by the targeted drug regimen in Section~\ref{cdcn}.

\subsection{Preliminaries}

\subsubsection{Ordinary Differential Equation System}
\label{odes}
Following~\citep{mcsharry2003dynamical}, we generate the EPK through coupled ODEs parameterized by heart-rate statistics and variability. The system produces the cardiac trajectory $(x(t_{\text{ode}}),y(t_{\text{ode}}),z(t_{\text{ode}}))$:
\begin{equation}
\label{ode}
\begin{aligned}
\frac{\mathrm{d}x}{\mathrm{d}t_{\text{ode}}}&=\alpha x-\omega y \equiv f_x(x,y;\eta), \\
\frac{\mathrm{d}y}{\mathrm{d}t_{\text{ode}}}&=\alpha y+\omega x \equiv f_y(x,y;\eta), \\
\frac{\mathrm{d}z}{\mathrm{d}t_{\text{ode}}}&=-\sum\limits_{i \in \{\text{P},\text{Q},\text{R},\text{S},\text{T}\}}a_i\Delta\theta_i \exp\left(-\frac{\Delta\theta_i^2}{2b_i^2}\right)-(z-z_0(t_{\text{ode}})) \\
& \equiv f_z(x,y,z,t_{\text{ode}};\eta),
\end{aligned}
\end{equation}
where $\eta=\{\omega,\theta_i,a_i,b_i\}$ collects the static parameters, $\alpha=1-\sqrt{x^2+y^2}$, $\Delta\theta_i=(\theta-\theta_i) \bmod 2\pi$, and $\theta=\mathrm{atan2}(y,x)$ (the four quadrant arctangent of the real parts of the elements of $x$ and $y$, with $-\pi \leq \mathrm{atan2}(y,x) \leq \pi$).
$\omega$ is the angular velocity of the trajectory as it moves around the limit cycle.
Baseline wander is injected through $z_0(t_{\text{ode}})=A\sin(2\pi f_2 t_{\text{ode}})$ with amplitude $A=0.15\text{mV}$ following~\citep{golany2020simgans}.
Baseline wander is a type of low-frequency noise ($0.05 - 3 \text{Hz}$ during stress testing)~\citep{sornmo2005bioelectrical}, caused by various factors: respiratory motion, patient movement, poor contact of electrode cables and ECG recording equipment, inadequate skin preparation at the electrode placement sites, and unclean electrodes~\citep{golany2020simgans}.
The $(x,y)$ components orbit a unit-radius limit cycle, while $z(t_{\text{ode}})$ traces the ECG morphology driven by $\theta$.
Each complete rotation of this circle corresponds to one heartbeat (or cardiac cycle).
Parameters $\theta_i$, $a_i$, and $b_i$ with $i\in\{\text{P}, \text{Q}, \text{R}, \text{S}, \text{T}\}$ fix the angular position, amplitude, and width of the PQRST complexes, respectively.
More details about the parameters can be found in Appendix~\ref{dodep}.

We integrate \eqref{ode} with the Euler method~\citep{atkinson1991introduction} using a fixed time step $\Delta t=1/f_s$ and $f_s=1000\text{Hz}$, which gives
\begin{displaymath}
\begin{aligned}
t_{\text{ode},l}&=l\Delta t, \\
x_{l+1}&=x_l+f_x(x_l,y_l;\eta)\Delta t, \\
y_{l+1}&=y_l+f_y(x_l,y_l;\eta)\Delta t, \\
z_{l+1}&=z_l+f_z(x_l,y_l,z_l,t_{\text{ode},l};\eta)\Delta t.
\end{aligned}
\end{displaymath}
Sampling $\{t_{\text{ode},l}\}_{l=1}^L$ yields $\mathbf{x}_{\text{ode}}=(z_l)_{l=1}^L$, \textit{i.e.}, the physics prior supplied to the diffusion model.

\subsubsection{Denoising Diffusion Probabilistic Model}
\label{ddpm}
Denoising Diffusion Probabilistic Models (DDPM)~\citep{ho2020denoising} are latent variable models that include a diffusion process where Gaussian noise is gradually added to the data until it reaches a Gaussian distribution, and a reverse process where the noise is converted back into samples.

The diffusion process gradually adds Gaussian noise to the data $\mathbf{x}_0 \sim q(\mathbf{x}_0)$ over $T$ steps, which is defined as a fixed Markov Chain:
\begin{displaymath}
\begin{aligned}
q(\mathbf{x}_{1:T}|\mathbf{x}_0)=&\prod \limits_{t=1}^T q(\mathbf{x}_t|\mathbf{x}_{t-1}),\\
q(\mathbf{x}_t|\mathbf{x}_{t-1})=&\mathcal{N}\left(\mathbf{x}_t; \sqrt{1-\beta_t}\mathbf{x}_{t-1},\beta_t\mathbf{I}\right),
\end{aligned}
\end{displaymath}
where $\mathbf{x}_1, ..., \mathbf{x}_T$ are the latent variables with the same dimensionality as $\mathbf{x}_0$, $\beta_1, ..., \beta_T$ is a variance schedule which ensures that $\mathbf{x}_T$ approximates a standard normal distribution for sufficiently large $T$, and $\mathcal{N}(\mathbf{x};\mu,\Sigma)$ is a Gaussian probability density function with parameters $\mu$ and $\Sigma$.
The sampling of $\mathbf{x}_t$ at an arbitrary step $t$ has the closed-form of $q(\mathbf{x}_t|\mathbf{x}_0)=\mathcal{N}(\mathbf{x}_t;\sqrt{\overline{\mathbf{\alpha}}_t}\mathbf{x}_0,(1-\overline{\mathbf{\alpha}}_t)\mathbf{I})$, where $\overline{\alpha}_t=\prod_{s=1}^t\alpha_s$ and $\alpha_s=1-\beta_s$.
Thus, $\mathbf{x}_t$ can be sampled in closed form as $\mathbf{x}_t=\sqrt{\overline{\alpha}_t}\mathbf{x}_0+\sqrt{1-\overline{\alpha}_t}\boldsymbol{\epsilon}$, where $\boldsymbol{\epsilon} \sim \mathcal{N}(\mathbf{0},\mathbf{I})$.

\begin{figure*}[tbp]
    \centering
    \includegraphics[width=1\textwidth]{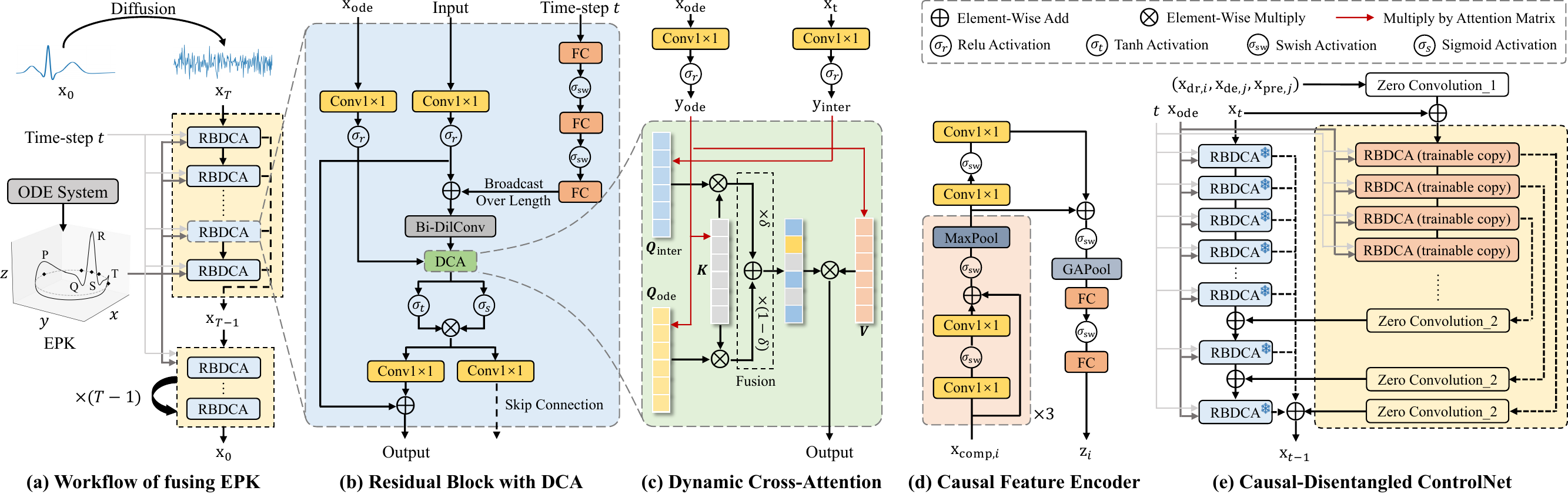}
    \caption{The architecture of MM-DADM. FC, MaxPool and GAPool denote fully connected, max pooling and global average pooling layers, respectively.}
    \label{1}
    \Description{This figure illustrates the MM-DADM model architecture, composed of five primary substructures. The leftmost part shows the macro workflow of MM-DADM (subfigure a); slightly to the left of center depicts the residual block details of RBDCA (subfigures b and c); slightly to the right of center presents the structural diagram of CFE (subfigure d); and the rightmost part shows CDC-Net (subfigure e), typically arranged in parallel with the main network as a side branch, injecting information unidirectionally into the backbone via arrows (representing zero convolution and fusion) to guide generation.}
\end{figure*}


The reverse process learns to invert the above diffusion process, restoring $\mathbf{x}_t$ to $\mathbf{x}_0$. Starting from pure Gaussian noise drawn from $p(\mathbf{x}_T) = \mathcal{N}(\mathbf{x}_T; \mathbf{0}, \mathbf{I})$, the reverse process is defined as follows:
\begin{displaymath}
\begin{aligned}
p_{\Theta_{\text{bb}}}(\mathbf{x}_{0:T})&=p(\mathbf{x}_T)\prod \limits_{t=1}^T p_{\Theta_{\text{bb}}}(\mathbf{x}_{t-1}|\mathbf{x}_t), \mathbf{x}_T \sim \mathcal{N}(\mathbf{0},\mathbf{I}),\\
p_{\Theta_{\text{bb}}}(\mathbf{x}_{t-1}|\mathbf{x}_t)&=\mathcal{N}(\mathbf{x}_{t-1};\boldsymbol{\mu}_{\Theta_{\text{bb}}}(\mathbf{x}_t,t),\phi_{\Theta_{\text{bb}}}(\mathbf{x}_t,t)\mathbf{I}),
\end{aligned}
\end{displaymath}
where $\Theta_{\text{bb}}$ denotes the parameters of the backbone (bb) model.
In DDPM's setting, $\boldsymbol{\mu}_{\Theta_{\text{bb}}}(\mathbf{x}_t,t)$ and $\phi_{\Theta_{\text{bb}}}(\mathbf{x}_t,t)$ are defined as
\begin{displaymath}
\boldsymbol{\mu}_{\Theta_{\text{bb}}}(\mathbf{x}_t,t)=\frac{1}{\alpha_t} \left(\mathbf{x}_t-\frac{\beta_t}{\sqrt{1-\alpha_t}}\boldsymbol{\epsilon}_{\Theta_{\text{bb}}}(\mathbf{x}_t,t)\right),
\end{displaymath}
\begin{displaymath}
\phi_{\Theta_{\text{bb}}}(\mathbf{x}_t,t)=\sqrt{\tilde{\beta}_t}, \tilde{\beta}_t=\left\{
\begin{aligned}
& \frac{1-\overline{\alpha}_{t-1}}{1-\overline{\alpha}_t}\beta_t, & t>1 \\
& \beta_1, & t=1 
\end{aligned}
\right.,
\end{displaymath}
where $\boldsymbol{\epsilon}_{\Theta_{\text{bb}}}(\cdot,\cdot)$ is a learnable denoising function used to predict the noise vector $\boldsymbol{\epsilon}$ added to $\mathbf{x}_t$.
The model is trained by optimizing the following simplified loss objective $\mathcal{L}_{\text{bb}}$:
\begin{displaymath}
\mathcal{L}_{\text{bb}}=\mathbb{E}_{t,\mathbf{x}_0,\epsilon}\left[ \Vert\boldsymbol{\epsilon}-\boldsymbol{\epsilon}_{\Theta_{\text{bb}}}(\mathbf{x}_t,t) \Vert^2\right],
\end{displaymath}
where $t$ is uniform between $1$ and $T$.

\subsection{Dynamic Cross-Attention Fusion}
\label{dcaf}
We fuse EPK into every denoising block via DCA (Figure~\ref{1}a).
Given that the DCA is a plug-and-play method, we specifically combine DCA with Residual Block~\citep{kongdiffwave} due to its effective framework for waveform generation, denoted by RBDCA.
In each RBDCA, as shown in Figure~\ref{1}b, we use a Bidirectional Dilated Convolution (Bi-DilConv) with kernel size $3$, where the dilation is doubled at each layer, \textit{i.e.}, $1,2,4,...,2^{n-1}$.
We use $N=30$ blocks grouped into $m=3$ stacks, so each stack contains $n=N/m=10$ RBDCA.
The output of the previous residual block is the input of the next one. Each block is also connected through a skip connection.

The EPK sequence $\mathbf{x}_{\text{ode}}$ is encoded into $\mathbf{y}_{\text{ode}}$, while the noisy latent $\mathbf{x}_t$ yields an intermediate feature $\mathbf{y}_{\text{inter}}$.
The EPK represents an idealized heartbeat, which may deviate from the complex pathological morphology induced by specific drugs.
To address this, DCA determines the fusion strength based on the feature discrepancy between $\mathbf{y}_{\text{ode}}$ and denoised latent features $\mathbf{y}_{\text{inter}}$ (Figure~\ref{1}c):
\begin{displaymath}
\text{DCA}:=\left\{ \left[ \delta \cdot \sigma\left(\frac{\mathbf{Q}_{\text{ode}}\mathbf{K}^\top}{\sqrt{d}}\right) \right] \oplus \left[ (1-\delta) \cdot \sigma\left( \frac{\mathbf{Q}_{\text{inter}}\mathbf{K}^\top}{\sqrt{d}} \right) \right] \right\} \mathbf{V},
\end{displaymath}
where $\mathbf{Q}_{\text{ode}}=\mathbf{W}^{q_1}\cdot \mathbf{y}_{\text{ode}}$,
$\mathbf{Q}_{\text{inter}}=\mathbf{W}^{q_2} \cdot \mathbf{y}_{\text{inter}}$,
$\mathbf{K}=\mathbf{W}^k\cdot\mathbf{y}_{\text{ode}}$,
$\mathbf{V}=\mathbf{W}^v\cdot\mathbf{y}_{\text{ode}}$,
$\mathbf{W}^{q_1}$, $\mathbf{W}^{q_2}$, $\mathbf{W}^k$ and $\mathbf{W}^v$ are learnable projection matrices~\citep{vaswani2017attention}.
$\sigma$ and $d$ denote the softmax function and the token dimension of $\mathbf{Q}_{\text{ode}}\mathbf{K}^\top$, respectively.
$\delta=\sigma_{\text{sw}}\left(\text{Conv1$\times$1}(|\mathbf{y}_{\text{ode}}-\mathbf{y}_{\text{inter}}|)\right)$ serves as a learnable tension meter, which measures the semantic distance between the rigid physical manifold and the evolving data manifold.

\subsection{Causal Feature Encoder Pre-Training}
\label{ept}
To accurately isolate the causal effect of each drug on the ECG morphology, we pre-train a CFE (Figure~\ref{1}d), denoted as $E$, and dynamically maintain a prototype for each drug.

First, we concatenate the pre-dose ECGs $\mathbf{x}_{\text{pre},i}$ and real post-dose ECGs $\mathbf{x}_{\text{post},i}$ along the channel dimension to construct a composite input tensor, \textit{i.e.}, $\mathbf{x}_{\text{comp}, i}=\text{Concat}(\mathbf{x}_{\text{pre},i},\mathbf{x}_{\text{post},i})$.
Here, instead of performing element-wise subtraction on the original signals, we adopt a concatenation operation. This allows subsequent steps to leverage the translation invariance of deep convolutional layers to dynamically align local features, thereby facilitating the separation of pure drug effect in the feature space. In contrast, direct element-wise subtraction inevitably leads to severe waveform mismatch, generating meaningless structural artifacts rather than genuinely isolating the drug residuals.
Then, the composite tensor is encoded by $E$ to obtain the feature vector $\mathbf{z}_{i} = E(\mathbf{x}_{\text{comp}, i})$.
To ensure the encoder extracts drug-specific representations rather than patient-specific noise, we employ contrastive learning. For a batch of composite inputs, the contrastive loss $\mathcal{L}_{\text{con}}$ is formulated to maximize the similarity of features from the same drug, while minimizing the similarity across different drugs:
\begin{displaymath}
\mathcal{L}_{\text{con}} = \sum_{i\in\mathcal{B}} \frac{-1}{| \mathcal{P}(i) |} \sum_{p \in \mathcal{P}(i)} \log \frac{\exp(\text{sim}(\mathbf{z}_{i}, \mathbf{z}_{p}) / \tau)}{\sum_{j \in \mathcal{B}} \exp(\text{sim}(\mathbf{z}_{i}, \mathbf{z}_{j}) / \tau)},
\end{displaymath}
where $\mathcal{B}$ is the indices of all samples in a batch, $\mathcal{P}(i)$ is the set of positive samples (same drug regimen) for anchor $i$, $\tau$ is a temperature hyperparameter, and $\text{sim}(\cdot, \cdot)$ denotes cosine similarity.

Concurrently, we maintain a dynamic prototype $\mathbf{p}_{i}$ for each drug using an Exponential Moving Average (EMA) mechanism~\citep{laine2016temporal}. These prototypes serve as the canonical representation of each drug's causal effect:
\begin{displaymath}
\mathbf{p}_i^{(e)} = \gamma \mathbf{p}_i^{(e-1)} + (1 - \gamma) \frac{1}{M_i} \sum_{j=1}^{M_i} \mathbf{z}_{j},
\end{displaymath}
where $e$ is the training epoch, $\gamma \in [0, 1)$ is the momentum parameter, and $M_i$ is the number of samples for drug $i$ in the current epoch.

\subsection{Causal-Disentangled ControlNet}
\label{cdcn}
Having obtained a robust encoder $E$ and established the drug prototype $\mathbf{p}_i$, we freeze the parameters of $E$ and the backbone model, and then train the CDC-Net.
CDC-Net adopts the model architecture of the original ControlNet~\citep{zhang2023adding}, which contains a trainable copy of the RBDCA denoted by $\mathrm{f}^{\prime}(\cdot,\cdot;\Theta^{\prime})$,
and two zero-initialized $1$D convolutions $\mathrm{z}_1(\cdot;\Theta_1)$ and $\mathrm{z}_2(\cdot;\Theta_2)$, as shown in Figure~\ref{1}e.
Zero-initialized convolution ensures that demographic and drug features do not influence ECG signal generation during the initial training phase.

To overcome the scarcity of real paired data, we introduce a counterfactual data augmentation strategy.
We synthetically recombine a patient's demographic covariates $\mathbf{x}_{\text{de},j}$ with unmatched drug-related variables $\mathbf{x}_{\text{dr},i}$ (\textit{i.e.}, $i\neq j$) to generate an expansive set of counterfactual training samples.
The forward pass for both real and counterfactual samples is the same, defined as
\begin{displaymath}
\mathbf{y}_{\text{cdc},i,j}= \mathrm{z}_2(\mathrm{f}^{\prime}(\mathbf{x}_{\text{ode}},\mathbf{x}_t \oplus \mathrm{z}_1(\text{Concat}(\mathbf{x}_{\text{dr},i},\mathbf{x}_{\text{de},j},\mathbf{x}_{\text{pre},j})))),
\end{displaymath}
where $\mathbf{y}_{\text{cdc}}$ is the output of CDC-Net.
Note that $i = j$ when the input consists of real samples, whereas $i \neq j$ when the input consists of counterfactual samples.
Then the model optimizes over both real and counterfactual samples.
For real samples: Since the ground-truth post-dose ECGs exist, the DCD-Net is optimized using the standard conditional denoising objective $\mathcal{L}_{\text{real}}$, defined as
\begin{displaymath}
    \mathcal{L}_{\text{real}} = \mathbb{E}\left[ \Vert\boldsymbol{\epsilon}-\boldsymbol{\epsilon}_{\Theta}(\mathbf{x}_t,t,\mathbf{x}_{\text{ode}},\mathbf{x}_{\text{dr}},\mathbf{x}_{\text{de}},\mathbf{x}_{\text{pre}}) \Vert^2\right],
\end{displaymath}
where $\boldsymbol{\epsilon}_{\Theta}$ with $\Theta=\{\Theta_{\text{bb}}, \Theta^{\prime},\Theta_1,\Theta_2\}$ denotes the full denoising network, explicitly conditioned on both the EPK, the demographics, and the drug information.
For counterfactual samples: Lacking a ground-truth post-dose ECG, we rely on the pre-trained prototypes for guidance.
We first construct the predicted composite tensor $\hat{\mathbf{x}}_{\text{comp},i,j} = \text{Concat}(\mathbf{x}_{\text{pre},j}, \mathbf{y}_{\text{cdc},i,j})$ and then extract its causal feature. We enforce a causal guidance loss $\mathcal{L}_{\text{cau}}$ to minimize the distance between the predicted drug effect and the corresponding prototype:
\begin{equation}
\label{caul}
\mathcal{L}_{\text{cau}} = \sum_{j\in\mathcal{B}} \sum_{i\neq j}\left(1 - \text{sim}(E(\hat{\mathbf{x}}_{\text{comp},i,j}), \mathbf{p}_i)\right).
\end{equation}
The total loss function $\mathcal{L}$ for training the CDC-Net is defined as
\begin{displaymath}
    \mathcal{L} = \mathcal{L}_{\text{real}} + \lambda \mathcal{L}_{\text{cau}},
\end{displaymath}
where $\lambda$ is a balancing coefficient. By injecting this causal guidance, CDC-Net is explicitly coerced to comprehend the intrinsic pharmacological mechanisms, ensuring that the synthesized variations are driven by the assigned drug rather than demographic correlations.

\section{Experiments}

In this section, we first outline the dataset and implementation details in Section~\ref{did}, and the clinical indicators used for assessment in Section~\ref{id}.
Next, we benchmark MM-DADM against SOTA generative models across multiple paradigms in Section~\ref{cogm}.
Then we assess the morphological fidelity and clinical realism of the synthesized ECGs in Section~\ref{egef}.
Furthermore, we demonstrate the practical utility of MM-DADM via a downstream drug classification task in Section~\ref{dte}.
Finally, we conduct ablation studies and qualitative visualizations to validate the effectiveness of our core architectural designs in Section~\ref{as} and Section~\ref{vi}, respectively.

\subsection{Dataset and Implementation Details}
\label{did}
We conduct all the experiments on two databases, ECGRDVQ~\citep{johannesen2014differentiating} and ECGDMMLD~\citep{johannesen2016late}.
They together provide $9{,}443$ $12$-lead ECGs from $44$ subjects across $8$ regimens: Dofetilide, Lidocaine+Dofetilide, Mexiletine+Dofetilide, Moxifloxacin+Diltiazem, Placebo, Quinidine, Ranolazine and Verapamil.
A stratified $4:1$ split preserves each drug's prevalence in the train and test sets.
More details of the dataset can be found in Appendix~\ref{dod}.

Following previous work~\citep{bondar2025flowecg}, we train our model to generate $8$ independent ECG leads, specifically the $6$ precordial leads plus leads I and aVF.
The remaining $4$ limb leads are reconstructed using established electrocardiographic relationships, \textit{i.e.},
$\text{II}=(2\cdot\text{aVF}+\text{I})/2$,
$\text{III}=\text{II}-\text{I}$,
$\text{aVL}=(\text{I}-\text{III})/2$, and
$\text{aVR}=-(\text{I}+\text{II})/2$.
This approach ensures that the generated ECGs maintain physiologically consistent lead relationships.

All models are trained on $8$ NVIDIA RTX 4090 GPUs. The training process is divided into three steps: (1) train a backbone DDPM that fuses EPK via DCA;
(2) train a CFE and maintain a prototype for each drug;
(3) train the CDC-Net while freezing the parameters of the backbone and the CFE.
We set $T=1000$ and the forward process variances to constants increasing linearly from $\beta_1=10^{-4}$ to $\beta_T=0.02$. We use the Adam optimizer~\citep{kingma2015adam} with a batch size of $16$ and a learning rate of $2\times10^{-4}$.
The momentum parameter $\gamma$ and the weight of the causal guidance loss $\lambda$ are set to $0.99$ and $0.1$, respectively.
We train all models for $50$ epochs.
During the validation phase, we generate the post-dose ECGs at specific time points by varying the drug information input to CDC-Net.

\subsection{Indicator Details}
\label{id}
We judge whether a generated ECG matches the real post-dose response by comparing their normal/abnormal status on $3$ clinical indicators: $\text{QT}_\text{c}$, PR, and $\text{T}_{\text{peak}}-\text{T}_{\text{end}}$ intervals.
A prediction counts as correct only when the generated indicator label agrees with its real counterpart.
The calculation rules and normal ranges of each indicator are as follows:
(1) $\text{QT}_\text{c}$ interval is the corrected QT interval, representing the time from ventricular depolarization to complete repolarization, adjusted for heart rate. Here, we use the Bazett formula for correction: $\text{QT}_\text{c}=\text{QT}/\sqrt{\text{RR}}$.
$\text{QT}_\text{c}\leq450\text{ms}$ in men and $\text{QT}_\text{c}\leq470\text{ms}$ in women are considered normal~\citep{gupta2007current}.
(2) PR interval represents the time it takes for the electrical signal to travel from the atria to the ventricles, including atrial depolarization and the delay at the atrioventricular node. $120 \text{ms} \leq \text{PR} \leq 200 \text{ms}$ is normal~\citep{kwok2016prolonged}.
(3) $\text{T}_{\text{peak}}-\text{T}_{\text{end}}$ interval represents the late phase of ventricular repolarization and reflects the dispersion of ventricular repolarization.
$80 \text{ms} \leq \text{T}_{\text{peak}}-\text{T}_{\text{end}} \leq 113 \text{ms}$ is considered normal~\citep{icli2015prognostic}.
The onset and offset points of the $3$ indicators are determined by $10$ cardiologists with at least $5$ years of experience.

\begin{table*}[tbp]
\caption{Comparison with other methods. The best performance is bold, and the second-best performance is underlined.}
\centering
\begin{tabular}{l|cc|cc|cc}
\toprule
\multicolumn{1}{c|}{}
& \multicolumn{2}{c|}{$\text{QT}_\text{c}$ Interval}   & \multicolumn{2}{c|}{PR Interval}   & \multicolumn{2}{c}{$\text{T}_{\text{peak}}-\text{T}_{\text{end}}$ Interval} \\
\multicolumn{1}{l|}{\multirow{-2}{*}{Methods}}
& Accuracy  & Recall  & Accuracy  & Recall  & Accuracy  & Recall  \\
\hline
\multicolumn{7}{l}{\cellcolor[HTML]{C0C0C0}\textit{Generative Adversarial Network}} \\
WGAN
& $74.96\%$   & $75.00\%$   & $74.96\%$   & $73.53\%$   & $78.53\%$   & $80.36\%$   \\
StyleGAN
& $73.59\%$   & $70.00\%$   & $75.13\%$   & $73.53\%$   & $77.34\%$   & $78.55\%$   \\
TTS-CGAN
& $70.53\%$   & $72.50\%$   & $70.53\%$   & $76.47\%$   & $83.48\%$   & $82.55\%$   \\
CECG‐GAN
& $73.59\%$   & $75.00\%$   & $72.91\%$   & $70.59\%$   & $78.19\%$   & $77.09\%$   \\
ECG ODE-GAN
& $63.20\%$   & $55.00\%$   & $62.69\%$   & $58.82\%$   & $74.11\%$   & $72.73\%$   \\
\hline
\multicolumn{7}{l}{\cellcolor[HTML]{C0C0C0}\textit{Diffusion Model}} \\
DiffECG
& $\underline{83.48\%}$   & $\underline{77.50\%}$   & $83.65\%$
& $82.35\%$   & $\underline{85.86\%}$   & $\underline{85.45\%}$   \\
SSSD-ECG
& $82.11\%$ & $75.00\%$ & $84.16\%$
& $\underline{85.29\%}$ & $84.50\%$ & $84.36\%$ \\
BioDiffusion
& $80.41\%$   & $72.50\%$   & $\underline{85.01\%}$   & $79.41\%$   & $83.82\%$   & $83.64\%$   \\
\hline
\multicolumn{7}{l}{\cellcolor[HTML]{C0C0C0}\textit{Variational Autoencoder}} \\
Auto-TTE
& $62.35\%$   & $60.00\%$   & $72.40\%$   & $67.65\%$   & $71.55\%$   & $73.45\%$   \\
\hline
\multicolumn{7}{l}{\cellcolor[HTML]{C0C0C0}\textit{Flow Matching Model}} \\
FlowECG
& $81.43\%$ & $75.00\%$ & $83.13\%$
& $79.41\%$ & $83.82\%$ & $82.91\%$ \\
\cmidrule{1-7}
MM-DADM (Ours)
& $\mathbf{90.29\%}$   & $\mathbf{87.50\%}$   & $\mathbf{91.14\%}$   & $\mathbf{91.18\%}$   & $\mathbf{92.16\%}$   & $\mathbf{93.45\%}$   \\
\bottomrule
\end{tabular}
\label{others}
\end{table*}

\subsection{Comparison with Other Generated Methods}
\label{cogm}
We compare MM-DADM with other methods, including five GANs, \textit{i.e.}, WGAN~\citep{arjovsky2017wasserstein}, StyleGAN~\citep{karras2019style}, ECG ODE-GAN~\citep{golany2021ecg}, TTS-CGAN~\citep{li2022tts}, and CECG‐GAN~\citep{yang2024data},
three DMs, \textit{i.e.}, DiffECG~\citep{neifar2023diffecg}, BioDiffusion~\citep{li2024biodiffusion}, and SSSD-ECG~\citep{alcaraz2023diffusion},
one VAE, \textit{i.e.}, Auto-TTE~\citep{chung2023text},
and one FMM, \textit{i.e.}, FlowECG~\citep{bondar2025flowecg}.
Because detecting drug-induced abnormalities is critical, we report both accuracy (ACC) and recall (REC).
High recall reflects reliable detection of pathological changes.

\subsubsection{Single Indicator Evaluation}
Table~\ref{others} shows that MM-DADM attains the top accuracy and recall on all $3$ indicators. For $\text{QT}_\text{c}$ interval, it surpasses the next best method by $6.81\%$ accuracy and $10\%$ recall.
Other diffusion models still outperform GAN-based baselines, underscoring the suitability of diffusion for ECG synthesis.
Despite also using ODE priors, ECG ODE-GAN lags behind MM-DADM, highlighting the importance of our fusion strategy.

\begin{table}[tbp]
\caption{Comparison in simulating composite drug reactions.}
\centering
\begin{tabular}{l|ccc}
\toprule
\multicolumn{1}{l|}{Methods} & \multicolumn{1}{c}{Lid+Dof} & \multicolumn{1}{c}{Mex+Dof} & \multicolumn{1}{c}{Mox+Dil} \\
\cmidrule{1-4}
WGAN                  & $39.22\%$ & $49.09\%$ & $62.75\%$\\
StyleGAN              & $31.37\%$ & $41.82\%$ & $70.59\%$\\
TTS-CGAN              & $50.98\%$ & $30.91\%$ & $60.78\%$\\
CECG-GAN              & $29.41\%$ & $43.64\%$ & $45.10\%$\\
ECG ODE-GAN           & $27.45\%$ & $32.73\%$ & $47.06\%$\\
DiffECG               & $43.14\%$ & $\mathbf{61.82\%}$ & $\underline{80.39\%}$\\
SSSD-ECG              & $47.06\%$ & $38.18\%$ & $\underline{80.39\%}$\\
BioDiffusion          & $\underline{62.75\%}$ & $\underline{60.00\%}$ & $64.71\%$\\
Auto-TTE              & $39.22\%$ & $36.36\%$ & $37.25\%$\\
FlowECG               & $47.06\%$ & $32.73\%$ & $78.43\%$\\
\cmidrule{1-4}
MM-DADM (Ours)           & $\mathbf{76.47\%}$ & $\mathbf{61.82\%}$ & $\mathbf{84.31\%}$\\
\bottomrule
\end{tabular}
\label{multi}
\end{table}

\subsubsection{Multi-Indicator Evaluation}

We further test composite drug effects for Lidocaine (Lid) + Dofetilide (Dof), Mexiletine (Mex) + Dofetilide, and Moxifloxacin (Mox) + Diltiazem (Dil), counting a sample as correct only when every affected indicator is reproduced.
The results in Table~\ref{multi} confirm that:
(1) MM-DADM is the most robust and stable model, achieving the highest overall accuracy across the tested settings (including tying for first place with DiffECG on the Mex+Dof combination);
(2) All methods experience a sizable accuracy drop versus the single-indicator task, indicating that simulating composite drug reactions remains challenging.

\subsection{Evaluation of Generated ECG Fidelity}
\label{egef}
Ensuring the high fidelity of generated post-dose ECGs is a crucial prerequisite for their application in virtual clinical trials. We evaluate their fidelity from two perspectives. First, we employ quantitative metrics to measure the distributional and structural similarities between generated and real ECGs. Second, we conduct an expert evaluation to assess the clinical realism of the generated signals.

\subsubsection{Quantitative Evaluation}
The Fr\'echet Inception Distance~\citep{heusel2017gans} is a traditional metric used for assessing the fidelity of generated images. Based on it, we adopt the Fr\'echet ResNet Distance (FRD) as the metric to evaluate the generated ECG fidelity following~\citep{hu2024personalized}.
Specifically, we utilize a 1D ResNet~\citep{hong2020holmes} pre-trained on Challenge 2017 data~\citep{clifford2017af} as the backbone for feature extraction from ECGs.
Then the output of the final fully connected layer is used to calculate the FRD.
In addition, we adopt Root Mean Squared Error (RMSE) to evaluate the similarity between the generated and real ECGs. Lower FRD and RMSE indicate that the distribution of generated ECGs is closer to that of real ECGs.
We compare the average FRD scores and RMSE of $12$-lead ECGs generated by different methods.
The results in Table~\ref{fid} demonstrate that our method achieves the optimal FRD score of $0.2431$ and an RMSE of $0.0127$.

\begin{table}[tbp]
\caption{Evaluation of the generated ECG fidelity.}
\centering
\begin{tabular}{l|cc}
\toprule
\multicolumn{1}{l|}{Methods} & \multicolumn{1}{c}{FRD ($\downarrow$)} & \multicolumn{1}{c}{RMSE ($\downarrow$)} \\
\cmidrule{1-3}
WGAN                  & $1.3466$                 & $0.2328$ \\
StyleGAN              & $1.7853$                 & $0.3021$ \\
TTS-CGAN               & $0.7880$                 & $0.0716$ \\
CECG-GAN              & $0.7336$                 & $0.0728$ \\
ECG ODE-GAN           & $2.5639$                 & $0.3764$ \\
DiffECG               & $0.5322$                 & $0.0412$ \\
SSSD-ECG              & $\underline{0.4719}$        & $\underline{0.0312}$ \\
BioDiffusion          & $0.4832$     & $0.0343$ \\
Auto-TTE              & $2.5234$                 & $0.4162$ \\
FlowECG               & $0.5164$        & $0.0383$ \\
\cmidrule{1-3}
MM-DADM (Ours)           & $\mathbf{0.2431}$        & $\mathbf{0.0127}$ \\
\bottomrule
\end{tabular}
\label{fid}
\end{table}

\begin{table}[tbp]
\caption{Accuracy of cardiologists in distinguishing between real and generated ECGs.}
\centering
\resizebox{\linewidth}{!}{
\begin{tabular}{l|cccccccc}
\toprule
\multicolumn{1}{l|}{Methods} & \multicolumn{1}{c}{V1} & \multicolumn{1}{c}{V2} & \multicolumn{1}{c}{V3} & \multicolumn{1}{c}{V4} & \multicolumn{1}{c}{V5} & \multicolumn{1}{c}{V6} & \multicolumn{1}{c}{I} & \multicolumn{1}{c}{aVF} \\
\cmidrule{1-9}
TTS-CGAN
& $61.0$   & $63.0$   & $\underline{59.0}$   & $64.0$   & $65.5$   & $63.0$   & $62.5$   & $62.0$   \\
DiffECG
& $\underline{60.0}$   & $62.0$   & $59.5$   & $\underline{60.5}$   & $\underline{63.5}$   & $63.5$   & $\underline{59.0}$   & $61.0$   \\
Auto-TTE
& $66.0$   & $66.0$   & $64.0$   & $64.5$   & $67.0$   & $70.5$   & $63.0$   & $63.0$   \\
FlowECG
& $63.5$   & $\underline{61.5}$   & $64.5$   & $61.0$   & $\underline{63.5}$   & $\underline{61.5}$   & $59.5$   & $\underline{60.0}$   \\
\cmidrule{1-9}
MM-DADM
& $\mathbf{53.0}$   & $\mathbf{55.0}$   & $\mathbf{53.5}$
& $\mathbf{54.0}$   & $\mathbf{51.5}$   & $\mathbf{54.0}$
& $\mathbf{53.5}$   & $\mathbf{54.0}$   \\
\bottomrule
\end{tabular}
}
\label{expe}
\end{table}

\subsubsection{Expert Evaluation}

To further validate the clinical realism of our method, we conduct a visual Turing test following established protocols for medical GMs. We select one representative method from each category to generate ECGs for leads V1-V6, lead I, and lead aVF. For each specific lead, each method generates $100$ ECGs, which are then randomly mixed with $100$ real ECGs. This mixed dataset of generated and real ECGs is evaluated by ten cardiologists, each with at least five years of clinical experience.
The cardiologists are tasked with blindly distinguishing whether each ECG trace is real or generated. In this evaluation paradigm, an ideal generative model should render the experts unable to distinguish real from fake, driving their classification accuracy toward random chance ($50\%$). Therefore, a lower classification accuracy (closer to $50\%$) indicates higher fidelity of the generated ECGs.

The experimental results, as shown in Table~\ref{expe}, reveal that MM-DADM consistently outperforms all baseline methods across all evaluated leads by successfully confusing the experts. Notably, the expert classification accuracy for MM-DADM hovers around $51.5\%$ to $55.0\%$. In contrast, existing GANs, DMs, VAEs, and FMMs exhibit significantly higher expert accuracy rates (often exceeding $60\%$ or even $70\%$), indicating that their generated samples contain distinguishable synthetic artifacts or lack natural physiological noise, making them relatively easy for cardiologists to identify.

\begin{table}[tbp]
\caption{Drug efficacy classification accuracy with varying numbers of generated ECGs.}
\centering
\begin{tabular}{l|cc}
\toprule
\multicolumn{1}{l|}{Dataset} & \multicolumn{1}{c}{Quinidine}   & \multicolumn{1}{c}{Verapamil}  \\
\cmidrule{1-3}
$1,000$ real ECGs        & $91.25\%$   & $91.75\%$   \\
\cmidrule{1-3}
$+100$ generated ECGs    & $94.25\%$   & $95.00\%$   \\
$+200$ generated ECGs    & $96.25\%$   & $95.75\%$   \\
$+500$ generated ECGs    & $97.75\%$   & $97.25\%$   \\
\bottomrule
\end{tabular}
\label{dct}
\end{table}

\subsection{Downstream Task Evaluation}
\label{dte}
We use the generated ECGs for data augmentation in downstream classification tasks to demonstrate their clinical authenticity and practical value.
Specifically, we design a binary classification experiment to identify the effects of two distinct medications: Quinidine and Verapamil.
All experiments are conducted using Lead II ECG signals. For each drug, we train four ResNet-1D models using the following datasets to classify whether the drug is administered:
(1) $1,000$ real ECGs;
(2) $1,000$ real $+$ $100$ generated ECGs;
(3) $1,000$ real $+$ $200$ generated ECGs;
(4) $1,000$ real $+$ $500$ generated ECGs.
The test set is identical for all models within each drug category, consisting of $200$ pre-dose and $200$ post-dose Lead II ECGs.

The results are shown in Table~\ref{dct}, demonstrating that our generated data helps improve the performance of classification models. More importantly, the cost of generating data is significantly lower than that of acquiring real data. Our method provides a viable solution to the problem of data scarcity in real medical scenarios.

\subsection{Ablation Study}
\label{as}
We conduct ablation studies on the fusion mechanisms of EPK, the causal-disentangled strategy (CDS), and the weight $\lambda$ of the causal guidance loss, respectively.
It is worth noting that CDS comprises two processes: pre-training of CFE and contrastive learning for causal disentanglement of demographics and drug information.

\begin{table}[tbp]
\caption{Ablation of fusing mechanisms.}
\centering
\begin{tabular}{l|ccccc}
\toprule
\multicolumn{1}{l|}{Methods}
& \multicolumn{1}{c}{NA}     & \multicolumn{1}{c}{FWA}   & \multicolumn{1}{c}{PSA}
& \multicolumn{1}{c}{EECA}   & \multicolumn{1}{c}{DCA (Ours)}   \\
\cmidrule{1-6}
FRD ($\downarrow$)
& $0.701$   & $0.632$   & $0.473$   & $0.433$   & $\mathbf{0.329}$   \\
\bottomrule
\end{tabular}
\label{afm}
\end{table}

\begin{table}[tbp]
\caption{Ablation of CDS.}
\centering
\begin{tabular}{l|cc|cc|cc}
\toprule
\multicolumn{1}{c|}{}
& \multicolumn{2}{c|}{$\text{QT}_\text{c}$}   & \multicolumn{2}{c|}{PR}   & \multicolumn{2}{c}{$\text{T}_{\text{peak}}-\text{T}_{\text{end}}$} \\
\multicolumn{1}{l|}{\multirow{-2}{*}{Methods}}
& ACC  & REC  & ACC  & REC  & ACC  & REC  \\
\cmidrule{1-7}
TTS-CGAN
& $70.53$   & $72.50$   & $70.53$   & $76.47$   & $83.48$   & $82.55$ \\
\quad w/ CDS
& $78.19$   & $75.00$   & $80.92$   & $79.41$   & $84.16$   & $83.64$ \\
\cmidrule{1-7}
DiffECG
& $83.48$   & $77.50$   & $83.65$   & $82.35$   & $85.86$   & $85.45$ \\
\quad w/ CDS
& $85.86$   & $80.00$   & $88.07$   & $85.29$   & $87.56$   & $86.91$ \\
\cmidrule{1-7}
Auto-TTE
& $62.35$   & $60.00$   & $72.40$   & $67.65$   & $71.55$   & $73.45$ \\
\quad w/ CDS
& $67.80$   & $67.50$   & $76.15$   & $73.53$   & $78.88$   & $80.36$ \\
\cmidrule{1-7}
FlowECG
& $81.43$   & $75.00$   & $83.13$   & $79.41$   & $83.82$   & $82.91$ \\
\quad w/ CDS
& $83.30$   & $80.00$   & $84.16$   & $82.35$   & $85.52$   & $84.73$ \\
\bottomrule
\end{tabular}
\label{acds}
\end{table}

\subsubsection{Fusing Mechanisms of EPK}
To validate the effectiveness of DCA, we compare the FRD of ECGs generated using DCA with those generated using other $4$ baselines: no attention (NA), fixed-weight attention (FWA), prob-sparse attention (PSA)~\citep{zhou2021informer}, and exogenous-to-endogenous cross-attention (EECA)~\citep{wang2024timexer}.
NA means without fusing EPK, and the $\delta$ of FWA is fixed at $0.5$.
For a fair comparison, the generative model with DCA does not have a CDC-Net.
The results are presented in Table~\ref{afm}, from which we observe that
(1) Generated ECGs integrated with EPK exhibit lower FRD scores compared to those without EPK fusion, proving the effectiveness of incorporating EPK.
(2) DCA outperforms other static fusion mechanisms, demonstrating that dynamically adjusting EPK's constraints is effective and particularly well-suited for our task.

\subsubsection{Causal-Disentangled Strategy}
\label{pcds}
To validate the effectiveness of the CDS, we select one representative method from each category and compare the performance of these methods with and without applying CDS.
The specific procedure is similar to the training process of MM-DADM: first, a CEF is pre-trained to obtain prototypes for each drug; then, counterfactual samples are constructed.
For real samples, the conditional loss is computed according to each method. While for counterfactual samples, contrastive loss is computed according to (\ref{caul}).
Finally, the model is optimized using both losses.
The experimental results are shown in Table~\ref{acds}, from which we observe that after adopting CDS, the accuracy and recall of each model in simulating drug effects improve, demonstrating that our CDS helps models better understand pharmacological mechanisms.

\begin{figure*}[tbp]
    \centering
    \includegraphics[width=1\textwidth]{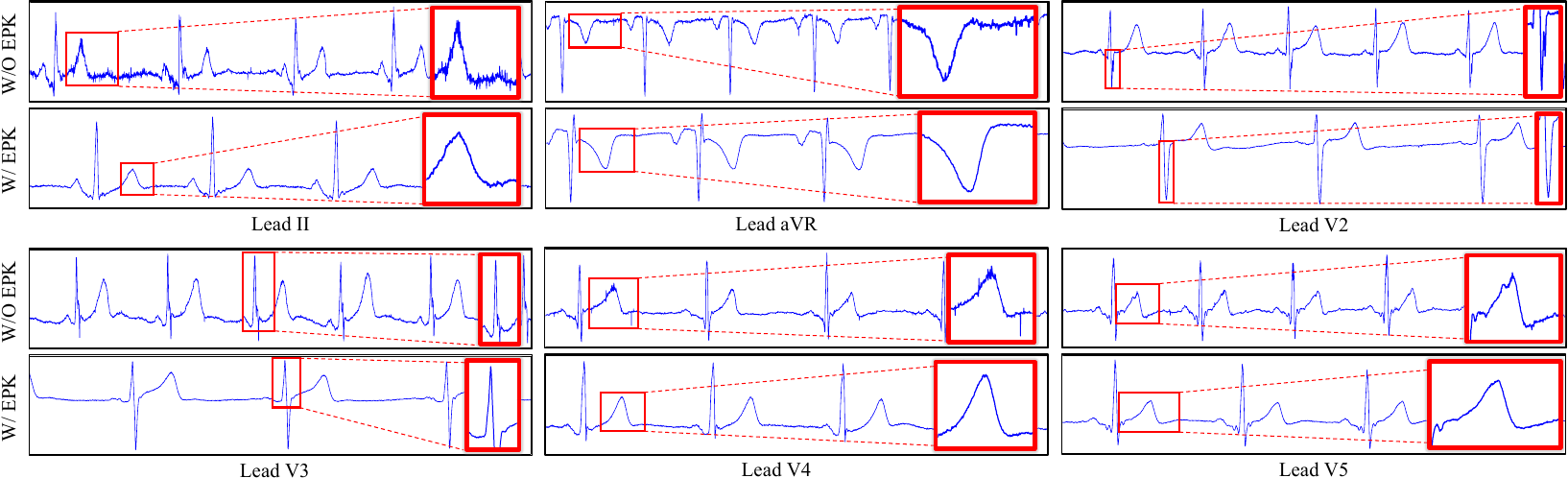}
    \caption{Visualization of ECG fidelity comparison.}
    \label{vis}
    \Description{This is a comparative diagram showing ECG signals generated with and without fused EPK.}
\end{figure*}


\begin{figure*}[tbp]
    \centering
    \includegraphics[width=1\textwidth]{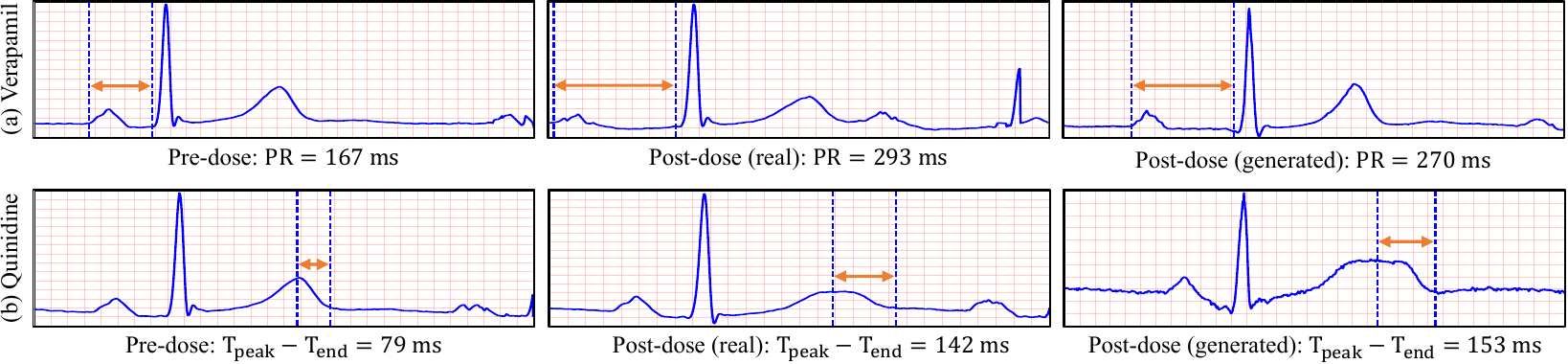}
    \caption{Visualization of Verapamil's and Quinidine's reaction on Lead II.}
    \label{vis2}
    \Description{This figure separately shows the ECG signals of subjects before and after administration of the drugs Verapamil (top) and Quinidine (bottom), as well as the post-administration ECG signals generated by MM-DADM. Specifically, the left side of each image displays the pre-dose ECGs, the center shows the actual post-dose signals, and the right side shows the generated post-dose signals.}
\end{figure*}


\begin{table}[tbp]
\caption{Ablation of $\lambda$.}
\centering
\begin{tabular}{l|cc|cc|cc}
\toprule
\multicolumn{1}{c|}{}
& \multicolumn{2}{c|}{$\text{QT}_\text{c}$}   & \multicolumn{2}{c|}{PR}   & \multicolumn{2}{c}{$\text{T}_{\text{peak}}-\text{T}_{\text{end}}$} \\
\multicolumn{1}{l|}{\multirow{-2}{*}{$\lambda$}}
& ACC  & REC  & ACC  & REC  & ACC  & REC  \\
\cmidrule{1-7}
$0.01$
& $83.48$ & $77.50$ & $82.79$ & $73.53$ & $85.52$ & $88.00$ \\
$0.1$
& $\mathbf{90.29}$   & $\mathbf{87.50}$
& $\mathbf{91.14}$   & $\mathbf{91.18}$
& $\mathbf{92.16}$   & $\mathbf{93.45}$   \\
$0.5$
& $82.79$ & $72.50$ & $82.11$ & $70.59$ & $83.99$ & $86.91$ \\
$1$
& $77.17$ & $62.50$ & $75.13$ & $61.77$ & $75.98$ & $77.09$ \\
\bottomrule
\end{tabular}
\label{asl}
\end{table}

\subsubsection{Hyper-Parameter $\lambda$}
We analyze the impact of different values of the hyperparameter $\lambda$ on the performance of CDC-Net.
The results are presented in Table~\ref{asl}, which show that the model achieves optimal performance when $\lambda=0.1$.

\subsection{Visualization}
\label{vi}
We present qualitative visualizations to demonstrate MM-DADM's capabilities, specifically highlighting the impact of EPK fusion and the simulation of pharmacological effects.


\subsubsection{Visualization of ECGs With and Without EPK}
Figure~\ref{vis} contrasts generated ECGs (leads II, aVR, and V2-V5) with and without DCA-facilitated EPK fusion. Incorporating EPK accurately captures fundamental heartbeat morphology and substantially reduces high-frequency noise, elevating clinical realism. Conversely, the baseline without EPK produces heavily distorted, noisy signals. Additional comparisons are provided in Appendix~\ref{avr}.


\subsubsection{Visualization of Drug Reactions}
Figure~\ref{vis2} demonstrates MM-DADM's precision in simulating drug-induced alterations. In Figure~\ref{vis2}a, Verapamil extends atrioventricular conduction time: the subject's PR interval increases from $167$ ms to an actual post-dose of $293$ ms. Our model closely mirrors this effect, predicting a $270$ ms PR interval.
Similarly, Figure~\ref{vis2}b illustrates Quinidine's prolongation of the $\text{T}_{\text{peak}}-\text{T}_{\text{end}}$ interval. The subject's true interval increases from a borderline $79$ ms to an abnormal $142$ ms, which MM-DADM successfully tracks by synthesizing a prolonged $153$ ms interval.
\section{Conclusions}

In this paper, we propose MM-DADM to simulate drug-induced ECG changes for virtual clinical trials. To overcome morphological rigidity and feature entanglement under limited clinical data, MM-DADM integrates a DCA module for adaptive physical prior fusion, alongside a CDC-Net to isolate pure pharmacological effects via counterfactual augmentation. Comprehensive evaluations on two public datasets demonstrate that MM-DADM outperforms $10$ SOTA GMs, improving simulation accuracy by at least $6.13\%$ and recall by $5.89\%$. Ultimately, MM-DADM provides a reliable, cost-effective computational reference for evaluating drug efficacy and cardiac safety, highlighting the potential of generative AI to reduce the scale and risk of physical trials safely.

\bibliographystyle{ACM-Reference-Format}
\bibliography{sample-base}

\appendix
\appendix

\section{Details of ODE Parameters}
\label{dodep}

\begin{figure}[htbp]
    \centering
    \includegraphics[width=0.6\linewidth]{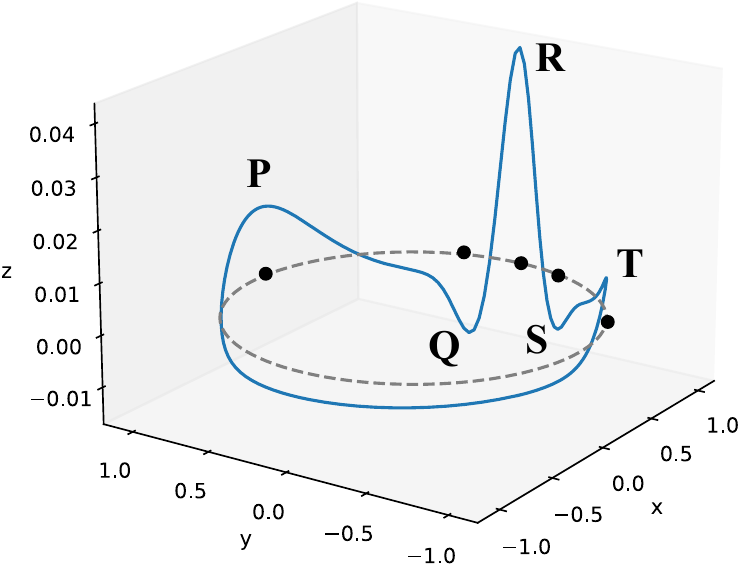}
    \caption{Visualization of the heartbeat in the 3D space. The dashed line reflects the limit cycle of unit radius, while the black points show the positions of the P, Q, R, S, and T waves.}
    \label{f1}
    \Description{This figure constructs a three-dimensional mathematical model that curls and maps the originally flat ECG signal into a 3D space, intuitively illustrating the continuous spatial evolution of the series of atrial/ventricular electrophysiological activities—P, Q, R, S, and T.}
\end{figure}

The ODE system is defined by $3$ parameter groups, including $\theta_i$: Phase positions governing wave event \underline{localization}; $a_i$: Amplitude parameters controlling the wave \underline{morphology}; $b_i$: Temporal scaling factors regulating the wave \underline{duration}, where $i\in\{\text{P}, \text{Q}, \text{R}, \text{S}, \text{T}\}$. To concretely demonstrate this parametric control, Figure~\ref{f1} visualizes a prototypical normal heartbeat with clinical reference values:
\begin{displaymath}
\begin{aligned}
(\theta_\text{P},\theta_\text{Q},\theta_\text{R},\theta_\text{S},\theta_\text{T})&=(-\frac{1}{3}\pi,-\frac{1}{12}\pi,0,\frac{1}{12}\pi,\frac{1}{2}\pi), \\
(a_\text{P},a_\text{Q},a_\text{R},a_\text{S},a_\text{T})&=(1.2,-5.0,30.0,-7.5,0.75), \\
(b_\text{P},b_\text{Q},b_\text{R},b_\text{S},b_\text{T})&=(0.25,0.1,0.1,0.1,0.4).
\end{aligned}
\end{displaymath}
We assign values to these parameters following~\citep{mcsharry2003dynamical}.

\section{Details of Dataset}
\label{dod}

\paragraph{ECG Recording Settings.}
The ECGs from the two databases are recorded at $1000$ Hz with an amplitude resolution of $2.5$ \textmu V.
For the ECGRDVQ database, triplicate $10$-second ECGs are extracted at $16$ predefined time points: $1$ point pre-dose ($-0.5$ h) and $15$ points post-dose ($0.5$, $1$, $1.5$, $2$, $2.5$, $3$, $3.5$, $4$, $5$, $6$, $7$, $8$, $12$, $14$, and $24$ h), during which the subjects were resting in a supine position for $10$ minutes.
For the ECGDMMLD database, triplicate $10$-second ECGs are extracted at $14$ predefined time points: $1$ point pre-dose ($-0.5$ h) and $13$ points post-dose ($1.5$, $2$, $2.5$, $3$, $6.5$, $7$, $7.5$, $8$, $12$, $12.5$, $13$, $13.5$, and $24$ h), during which the subjects were resting in a supine position for $10$ minutes.
We randomly select one ECG signal from the $3$ records at each time point for experiments.
We combine the pre- and post-dose ECGs with the corresponding demographics and drug information into a real training sample. This means that for each subject, $N$ samples can be generated, where $N$ represents the number of time points at which post-dose ECGs are recorded.

\paragraph{Demographic Information.}
Demographic information includes sex, age, height, weight, baseline systolic blood pressure, baseline diastolic blood pressure, race, and ethnicity.

\paragraph{Drug Information.}
Drug information includes sequence of treatments, visit code (PERIOD-X-Dosing refers to the X-th dosing), treatment (types of drug regimens), and nominal time-point. The nominal time point is the time to record post-dose ECGs.
The drugs in the datasets cause the following cardiac reactions:
\begin{itemize}[leftmargin=*]
    \item \textbf{Dofetilide} is mainly used to maintain the sinus rhythm in patients with atrial fibrillation and atrial flutter. It selectively blocks the delayed rectifier potassium current, prolonging the repolarization process of myocardial cells.
    \item \textbf{Lidocaine} and \textbf{Mexiletine} are mainly used for the treatment of ventricular arrhythmia. They may shorten the QT interval.
    \item \textbf{Moxifloxacin} is mainly used for the treatment of bacterial infections rather than directly targeting heart diseases. It may cause a prolongation of the QT interval.
    \item \textbf{Diltiazem} is mainly used for the treatment of arrhythmia, angina pectoris, and hypertension. It prolongs PR interval.
    \item \textbf{Placebo} is a substance or treatment that looks like real medicine but contains no active medical ingredients. Despite having no therapeutic properties, placebos can sometimes produce real physical or psychological benefits, known as the placebo effect.
    \item \textbf{Quinidine} is mainly used for the treatment of arrhythmia. It prolongs the QT interval and the width of the QRS complex.
    \item \textbf{Ranolazine} is mainly used for the treatment of chronic angina pectoris. It has a certain impact on the repolarization process of myocardial cells, which may lead to a moderate prolongation of the QT interval.
    \item \textbf{Verapamil} is mainly used for the treatment of arrhythmia, angina pectoris, and hypertension. It inhibits the influx of calcium ions, slowing down the conduction speed of the sinoatrial node and atrioventricular node, and prolonging the time it takes for the atrial impulse to be conducted to the ventricle.
\end{itemize}

\section{Additional Visualization Results}
\label{avr}

\begin{figure*}[tbp]
    \centering
    \includegraphics[width=1\textwidth]{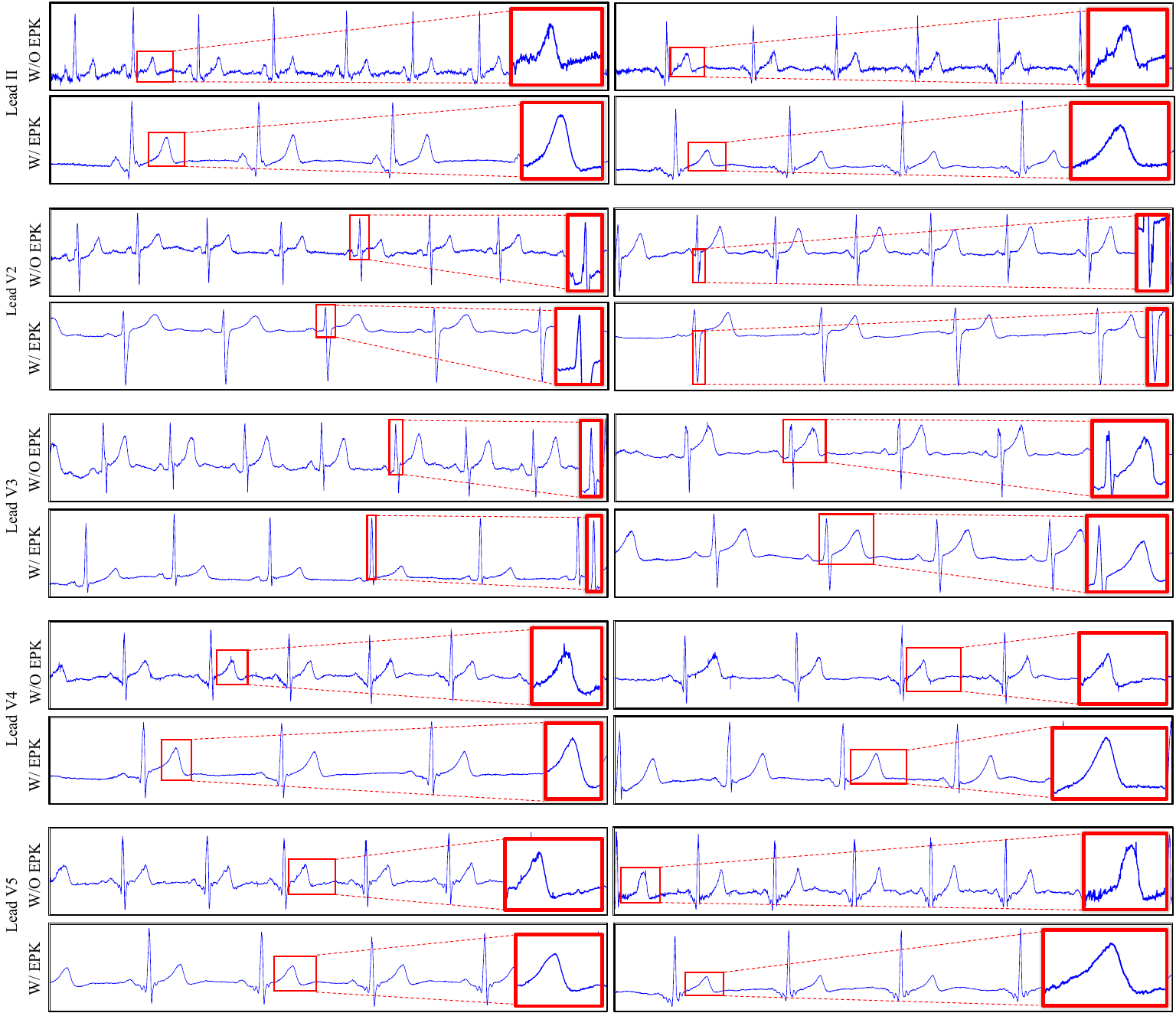}
    \caption{Additional visualization results of generated ECGs with and without EPK.}
    \label{app_epk}
    \Description{This is a comparative diagram showing ECG signals generated with and without fused EPK.}
\end{figure*}

Figure~\ref{app_epk} presents additional visualizations w/ and w/o EPK, from which we can observe that the introduction of EPK makes the generation of the ECG signal more realistic in detail. In the absence of EPK, the generated ECG signal contains a lot of noise.

\end{document}